\documentclass[runningheads]{llncs}
\usepackage[T1]{fontenc}
\usepackage{graphicx}
\usepackage{amsmath}
\usepackage{amssymb}
\usepackage{booktabs}
\usepackage{array}
\usepackage{algorithm}
\usepackage{algpseudocode}
\usepackage{url}
\newcolumntype{P}[1]{>{\raggedright\arraybackslash}p{#1}}

\begin{document}

\title{Three-Phase Scribble-Adaptive Curriculum Learning for autoPETV Grand Challenge}
\titlerunning{Curriculum Learning for autoPETV Grand Challenge}


\author{Libo Zhang \and Yue Ning}
\authorrunning{L. Zhang and Y. Ning}
\institute{Stevens Institute of Technology\\
\email{lzhang6@stevens.edu, Yue.Ning@stevens.edu}} 

\maketitle

\begin{abstract}
This report describes Libo Zhang's algorithmic solution to autoPETV Grand Challenge on interactive lesion segmentation in whole-body PET/CT. Interaction is encoded as two additional input channels that rasterize the accumulated foreground and background scribbles, and a residual-encoder U-Net of about 140 million parameters is trained with a three-phase curriculum over 4000 epochs: the network first learns fully automatic segmentation with silent interaction channels, then observes ground-truth-derived scribbles under randomly sampled visibility modes, and finally adapts to its own mistakes through online simulation of up to five error-driven correction steps. Training draws on 1811 autoPET and DeepPSMA studies, and the submission ensembles the best and final checkpoints of five folds by logit averaging. In interactive five-fold cross-validation with six interaction steps, the final checkpoints reach a mean AUC-Dice of 3.836 and a mean AUC-DMM of 3.869, improving monotonically in every fold, with roughly half of the total gain delivered by the first corrective scribble. Our code and trained model checkpoints are available on \url{https://github.com/Libo1023/autoPETV-Curriculum}.

\end{abstract}

\section{Introduction}
The autoPET challenge series has progressed from fully automated single-tracer lesion segmentation to multitracer, multicenter generalization and, most recently, to human-in-the-loop interaction \cite{gatidis2024results,dexl2026autopet,dexl2026autopet3,megnechoudja2026autopet}. Its fifth edition \cite{autopetv2026website,kuestner2026autopet} benchmarks interactive segmentation of tracer-avid tumor lesions in whole-body PET/CT: an algorithm produces an initial mask and then refines it as sparse corrective scribbles arrive over several iterations, and performance is scored by the area under the per-step Dice and detection-matching-metric curves (AUC-Dice and AUC-DMM), each contributing one half of the final ranking.

This report documents our submission. We model interaction as two extra input channels carrying the accumulated foreground and background scribbles, and align training with the inference-time correction process through a three-phase curriculum that culminates in online, error-driven multi-step scribble simulation. We merge the DeepPSMA cohort into the training split of every fold, and at test time we ensemble the best and final checkpoints of all five folds.

\section{Methods}

\subsection{Data}
\subsubsection{Training data}
We train on 1811 PET/CT studies: the 1014 FDG studies (University Hospital T\"ubingen) and 597 PSMA studies (LMU University Hospital Munich) of the official autoPET database \cite{gatidis2022whole,dexl2026autopet3}, plus the 200 studies (100 patients, paired PSMA and FDG scans) of the DeepPSMA challenge acquired at the Peter MacCallum Cancer Centre \cite{jackson2025psmafdg,buteau2022psma}. All 200 DeepPSMA studies are added to the training split of every fold while the validation folds remain identical to the official reference split. Confining DeepPSMA to the training side serves four purposes: it turns the incorporation of the auxiliary cohort into a controlled intervention that remains comparable to the challenge baselines; it avoids evaluating against threshold-derived total tumor burden labels, whose boundary convention differs from the free-hand expert annotations of the validation set; it preserves the complete center, scanner, and tracer diversity of a cohort that constitutes merely 11 percent of the corpus; and it rules out patient-level leakage between the paired FDG and PSMA studies of each patient.

\subsubsection{Validation data}
Validation uses the five folds of the official reference split, each containing 322 or 323 original autoPET studies and no DeepPSMA data.

\subsection{Data pre-processing}
\label{sec:preproc}
DeepPSMA CT volumes are resampled linearly to their PET grids and the total-tumor-burden masks are used as tumor labels. Every case is represented as a four-channel volume comprising CT, PET (SUV), and the two scribble channels introduced in Sect.~\ref{sec:model}. For training, scribble channels are pre-generated from the ground truth with three complementary strategies: a centerline strategy that traces the longest path of the lesion skeleton, a random strategy that connects two lesion voxels by a straight line, and a boundary strategy that walks along the inner lesion contour. Up to five lesion components are annotated per case, background scribbles are drawn from a two-voxel dilation ring surrounding the lesions, and all heatmaps are kept binary. To prevent any single scribble style from dominating the learned interaction behavior, exactly one strategy is assigned to every case in a round-robin fashion, which partitions the corpus into three essentially equal subsets of 604, 604, and 603 cases and thereby guarantees that each strategy governs one third of the training distribution. Preprocessing then resamples all channels to the median spacing of $3.0 \times 2.04 \times 2.04$ mm (third-order spline for images, linear for labels), normalizes CT with global foreground statistics (clipping to $[-822.5, 1127.0]$, followed by z-scoring with mean 126.7 and standard deviation 286.7), and finally applies per-image z-scoring to the PET channel and to both interaction channels.

\subsection{Algorithm/model}
\label{sec:model}
Let $x^{\mathrm{ct}}, x^{\mathrm{pet}} : \Omega \to \mathbb{R}$ denote the co-registered volumes on the voxel domain $\Omega$, let $y : \Omega \to \{0,1\}$ be the ground-truth lesion mask, and let $S_t^{+}, S_t^{-} \subset \Omega$ be the sets of foreground and background scribble voxels accumulated up to interaction round $t$, with $S_0^{\pm} = \emptyset$. The interaction state is rasterized into two binary channels and the segmentation at round $t$ is the stateless prediction
\begin{equation}
\hat{y}_t = \operatorname*{arg\,max}\, f_{\theta}\bigl(x^{\mathrm{ct}}, x^{\mathrm{pet}}, h_t^{+}, h_t^{-}\bigr), \qquad h_t^{\pm}(v) = \mathbf{1}\bigl[v \in S_t^{\pm}\bigr],
\label{eq:interactive}
\end{equation}
where $v \in \Omega$ indexes voxels, $f_{\theta}$ is the segmentation network with parameters $\theta$, the arg-max is taken over the two classes, and $\mathbf{1}[\cdot]$ denotes the indicator function; the entire interaction history thus enters the network exclusively through the heatmap channels $h_t^{\pm}$. The backbone realizing $f_{\theta}$ is the residual-encoder U-Net of nnU-Net \cite{isensee2021nnu,isensee2024nnu}, planned under a deliberately raised compute budget of 40 GB; since the medium, large, and extra-large presets share a single architecture template and differ only in this budget, the resulting configuration is of extra-large scale, with 7 stages of 32 to 320 features, encoder depths $(1,3,4,6,6,6,6)$, one decoder convolution per stage, deep supervision, patch size $192 \times 256 \times 256$, batch size 2, and about 140 million (M) parameters. The size of each model checkpoint is approximately 1.1 gigabytes (GB). Training aligns the distribution of the channels $h^{\pm}$ with the states encountered at inference through the three-phase curriculum summarized in Algorithm~\ref{alg:curriculum}. In Phase A (epochs 0 to 200) both channels are suppressed, $h^{\pm} \equiv 0$, and the network acquires purely automatic segmentation. In Phase B (epochs 200 to 3200) each sample draws a visibility mode $m$ from a categorical distribution over $\{\text{none},\, \text{both},\, \text{foreground only},\, \text{background only}\}$ with probabilities 0.20, 0.25, 0.35, and 0.20, and the pre-generated ground-truth scribble channels $(g^{+}, g^{-})$ of Sect.~\ref{sec:preproc} are masked accordingly, exposing the network to complete, one-sided, and absent guidance. In Phase C (epochs 3200 to 4000) the channels remain suppressed with probability 0.20; otherwise a single gradient-free forward pass with $h^{\pm} \equiv 0$ yields the automatic prediction $\hat{y}_0$ of Eq.~\eqref{eq:interactive}, from which the false-negative and false-positive sets
\begin{equation}
\mathcal{E}^{+} = \{v : y(v) = 1,\ \hat{y}_0(v) = 0\}, \qquad \mathcal{E}^{-} = \{v : y(v) = 0,\ \hat{y}_0(v) = 1\}
\label{eq:errors}
\end{equation}
are computed against the ground truth $y$. A correction depth $K$ is sampled uniformly from $\{1,\dots,5\}$ and $K$ steps are unrolled analytically: each step places one scribble, drawn with a randomly chosen strategy, on the largest axial connected component of the currently larger error set, and removes that component from the set so that successive steps attend to new regions, in the spirit of simulated-interaction training \cite{wang2018deepigeos,sofiiuk2022reviving,marinov2024deep}. The union of the placed scribbles is rasterized as in Eq.~\eqref{eq:interactive}, z-scored, and written into the interaction channels before the gradient-carrying pass. The cost per sample is thus two forward passes irrespective of $K$, which raised the epoch time only from roughly 250 s to 550 s and kept the 4000-epoch schedule feasible for all five folds. At inference the accumulated lesion clicks of the challenge interface \cite{hadlich2024sliding} are converted into the binary heatmaps of Eq.~\eqref{eq:interactive} on the PET grid, and the stateless network is re-run once per interaction step of the evaluation loop.

\begin{algorithm}[t]
\caption{Curriculum construction of the interaction channels $(h^{+}, h^{-})$ for one training sample at epoch $e$. Here $(g^{+}, g^{-})$ are the pre-generated ground-truth scribble channels, $\sigma \in \{+,-\}$ marks the side receiving the scribble, $s$ is the drawn scribble, and $C$ is its axial connected component in the error set.}
\label{alg:curriculum}
\begin{algorithmic}[1]
\Require $x^{\mathrm{ct}}, x^{\mathrm{pet}}, y$, epoch $e$, network $f_{\theta}$, channels $(g^{+}, g^{-})$
\Ensure interaction channels $(h^{+}, h^{-})$
\If{$e < 200$} \Comment{Phase A: automatic segmentation}
    \State \Return $(0, 0)$
\ElsIf{$e < 3200$} \Comment{Phase B: ground-truth-derived guidance}
    \State draw mode $m \sim \mathrm{Cat}(0.20, 0.25, 0.35, 0.20)$ over $\{$none, both, fg only, bg only$\}$
    \State \Return $(g^{+}, g^{-})$ masked according to $m$
\Else \Comment{Phase C: error-driven self-adaptation}
    \State with probability $0.20$: \Return $(0, 0)$
    \State $\hat{y}_0 \gets \operatorname*{arg\,max} f_{\theta}(x^{\mathrm{ct}}, x^{\mathrm{pet}}, 0, 0)$ \Comment{gradient-free pass}
    \State compute error sets $\mathcal{E}^{+}, \mathcal{E}^{-}$ by Eq.~\eqref{eq:errors}; \quad $S^{+} \gets \emptyset$; $S^{-} \gets \emptyset$
    \State draw correction depth $K \sim \mathcal{U}\{1,\dots,5\}$
    \For{$k = 1, \dots, K$}
        \State pick side $\sigma \in \{+,-\}$ whose candidate scribble is larger
        \State draw scribble $s$ (random strategy) on the largest axial component $C \subseteq \mathcal{E}^{\sigma}$
        \State $S^{\sigma} \gets S^{\sigma} \cup s$; \quad $\mathcal{E}^{\sigma} \gets \mathcal{E}^{\sigma} \setminus C$ \Comment{consume the corrected region}
        \If{$\mathcal{E}^{+} = \emptyset$ and $\mathcal{E}^{-} = \emptyset$} \textbf{break}
        \EndIf
    \EndFor
    \State \Return z-scored rasterizations of $(S^{+}, S^{-})$ \Comment{as in Eq.~\eqref{eq:interactive}}
\EndIf
\end{algorithmic}
\end{algorithm}

\subsection{Data post-processing}
After 5-fold training, we collect the five best checkpoints and the five final checkpoints for 10-fold ensemble prediction based on Eq.~\eqref{eq:interactive}. 

\subsection{Training and test parameters}
The networks are trained for $E = 4000$ epochs of 250 iterations with SGD (Nesterov momentum 0.99, weight decay $3 \times 10^{-5}$) under the polynomial schedule $\eta_e = \eta_0 (1 - e/E)^{0.9}$, where $e$ is the epoch index and $\eta_0 = 5 \times 10^{-4}$ the initial learning rate, minimizing the deeply supervised compound objective
\begin{equation}
\mathcal{L} = \sum_{d} w_d \bigl( \mathcal{L}_{\mathrm{CE}}^{(d)} + \mathcal{L}_{\mathrm{Dice}}^{(d)} \bigr), \qquad w_d \propto 2^{-d},
\label{eq:loss}
\end{equation}
where $d$ indexes the decoder resolutions, $\mathcal{L}_{\mathrm{CE}}^{(d)}$ and $\mathcal{L}_{\mathrm{Dice}}^{(d)}$ are the cross-entropy and soft Dice terms at resolution $d$, and the weights $w_d$ are normalized to sum to one with the lowest resolution excluded. Patches are sampled with 33 percent foreground oversampling, and the standard nnU-Net augmentation pipeline (rotation $\pm 30^{\circ}$, scaling from 0.7 to 1.4, Gaussian noise and blur, brightness and contrast from 0.75 to 1.25, simulated low resolution, gamma, mirroring on all axes) acts on all four channels on top of the curriculum-driven channel manipulation. For each fold we retain two checkpoints, the best by an exponential moving average of the online pseudo-Dice and the final at epoch 4000. Each fold occupies one NVIDIA L40S (48 GB) for about two weeks under PyTorch 2.6.0 with CUDA 12.4 and nnU-Net v2.6.0. At test time, sliding-window inference (half-patch overlap, Gaussian weighting, no test-time augmentation) is applied by all ten checkpoints, whose logits $z_i$, indexed by checkpoint $i$, are averaged into $\bar{z} = \frac{1}{10} \sum_{i=1}^{10} z_i$ before the arg-max; the submitted container executes one interaction step per invocation within the 20-minute limit. Table~\ref{tab1} consolidates these algorithm details in the format requested by the challenge.

\begin{table}[!ht]
\caption{Algorithm details.}\label{tab1}
\centering
\small
\setlength{\tabcolsep}{4pt}
\setlength{\aboverulesep}{0.2ex}
\setlength{\belowrulesep}{0.35ex}
\begin{tabular}{@{}P{0.30\textwidth}P{0.67\textwidth}@{}}
\toprule
\textbf{Team name} & LiboZhang \\
\textbf{Algorithm name} & autoPETV-Curriculum \\
\midrule
\textbf{Data pre-processing} & DeepPSMA CT linearly resampled to the PET grid; all channels resampled to the median spacing of $3.0 \times 2.04 \times 2.04$ mm (spline order 3 for images, linear for labels); CT clipped to $[-822.5, 1127.0]$ and z-scored (global mean 126.7, standard deviation 286.7); PET and both interaction channels z-scored per image; scribbles rasterized as two binary heatmaps \\
\textbf{Data post-processing} & Arg-max of the ensemble-averaged logits \\
\textbf{Training data augmentation} & Default nnU-Net pipeline on all four channels (rotation $\pm 30^{\circ}$, scaling 0.7 to 1.4, Gaussian noise and blur, brightness and contrast 0.75 to 1.25, simulated low resolution, gamma, mirroring on all axes); three-phase interaction curriculum of Sect.~\ref{sec:model} \\
\midrule
\textbf{Standardized framework} & nnU-Net v2 (2.6.0) on PyTorch 2.6.0 with CUDA 12.4 \\
\textbf{Network architecture} & 3D residual-encoder U-Net at a 40 GB compute budget: 7 stages with 32 to 320 features, encoder depths $(1,3,4,6,6,6,6)$, deep supervision, about 140M parameters, each checkpoint's size is roughly 1.1 GB \\
\textbf{Loss} & Dice plus cross-entropy with equal weights under deep supervision, Eq.~\eqref{eq:loss} \\
\textbf{Training data} & 1014 FDG and 597 PSMA studies of autoPET plus 200 DeepPSMA studies (training splits only); 1811 in total \\
\textbf{Data/model dimensionality and size} & 3D; patch size $192 \times 256 \times 256$, batch size 2 \\
\textbf{Pre-trained models} & None; all folds trained from scratch \\
\textbf{Training hardware} & NVIDIA L40S GPU (48 GB) \\
\midrule
\textbf{Test-time augmentation} & None (mirroring disabled) \\
\textbf{Ensembling} & Ten checkpoints (best and final of each fold), combined by voxel-wise logit averaging \\
\bottomrule
\end{tabular}
\end{table}

\clearpage
\section{Results}
Every fold and both checkpoints are evaluated with an interactive protocol mirroring the simulated regime of the challenge: six steps per case (an automatic prediction followed by five corrections), one scribble added per step on the currently larger error class, scribble strategies assigned to cases in the same round-robin fashion as during training so that each strategy covers one third of every validation fold, and lesion-level detection scored by matching 18-connected components at an intersection-over-union threshold of 0.1 \cite{kofler2023panoptica}. Writing $M_t$ for the value of the per-case metric $M \in \{\mathrm{Dice}, \mathrm{DMM}\}$ after interaction step $t \in \{0,\dots,5\}$, the challenge score is the trapezoidal area
\begin{equation}
\text{AUC-}M = \sum_{t=0}^{4} \tfrac{1}{2}\bigl(M_t + M_{t+1}\bigr),
\label{eq:auc}
\end{equation}
and lesion-free cases (107 to 123 per fold) are excluded by the metric definition. Table~\ref{tab:cv} reports every fold overall and per tracer, Fig.~\ref{fig:curves} shows the per-step curves, and the training curves of all folds are provided in Appendix~A. All ten runs improve monotonically over the interaction steps, and the first corrective scribble alone contributes about 55 percent of the total Dice gain (five-fold mean of the final checkpoints: 0.657 to 0.742 after one step, 0.810 after five). The final checkpoint is better in nine of ten fold-metric comparisons (mean advantage $+0.029$ AUC-Dice and $+0.042$ AUC-DMM), yet the best checkpoint wins both metrics on 20 percent of cases and provides the single strongest run (fold 4, AUC-Dice 3.898), which motivates ensembling both checkpoints of every fold. At every step and in every fold, Dice is higher on FDG while DMM is higher on PSMA, reflecting the larger individual lesions of the FDG cohort and the higher lesion counts of the PSMA cohort. The per-step curves are nearly identical across the three scribble strategies (final-checkpoint AUC-Dice between 3.74 and 3.93 per strategy within folds), a result indicating that the learned interaction behavior remains robust to the style of the scribbles simulated at test time.

\begin{figure}[p]
\centering
\includegraphics[scale=0.75]{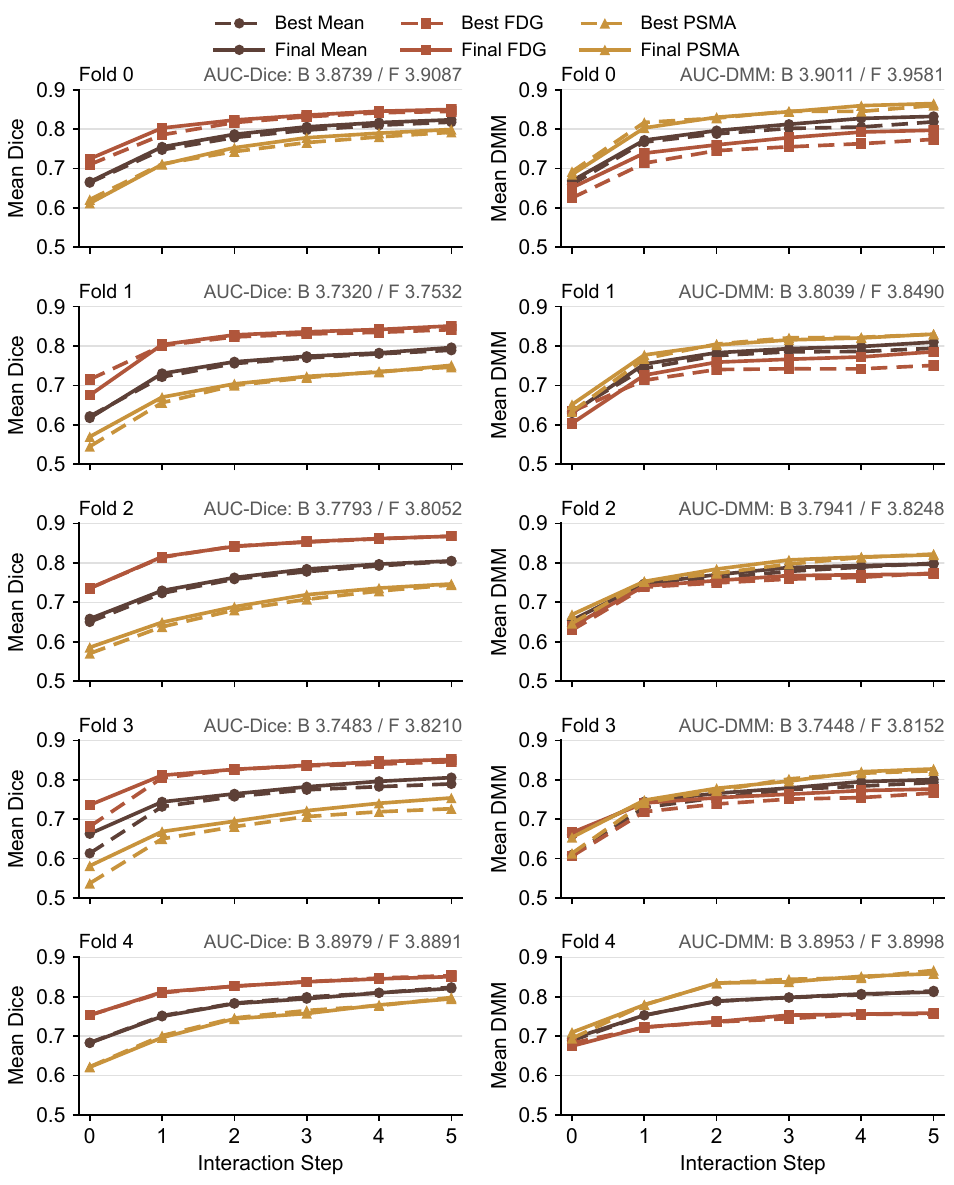}
\caption{Per-step mean Dice (left column) and DMM (right column) over the six-step interactive evaluation across five folds. Each panel overlays the best (dashed) and final (solid) checkpoints for all lesion-positive cases (circles) and for the FDG (squares) and PSMA (triangles) subsets; the AUC-Dice and AUC-DMM of both checkpoints are annotated per panel. Metrics are the areas under these curves following Eq.~\eqref{eq:auc}.}
\label{fig:curves}
\end{figure}

\begin{table}[t]
\caption{Interactive five-fold cross-validation of the best (B) and final (F) checkpoints, overall and per tracer. $n$ is the number of lesion-positive validation cases of each fold, and the better checkpoint per fold and column is highlighted in bold.}
\label{tab:cv}
\centering
\footnotesize
\setlength{\tabcolsep}{5pt}
\begin{tabular}{llcccccc}
\toprule
 & & \multicolumn{3}{c}{AUC-Dice} & \multicolumn{3}{c}{AUC-DMM} \\
\cmidrule(lr){3-5} \cmidrule(lr){6-8}
Fold ($n$) & Ckpt & All & FDG & PSMA & All & FDG & PSMA \\
\midrule
0 (207) & B & 3.8739 & 4.0525 & 3.7070 & 3.9011 & 3.6764 & 4.1110 \\
        & F & \textbf{3.9087} & \textbf{4.0936} & \textbf{3.7359} & \textbf{3.9581} & \textbf{3.7940} & \textbf{4.1114} \\
\addlinespace
1 (215) & B & 3.7320 & 4.0672 & 3.4565 & 3.8039 & 3.6299 & 3.9470 \\
        & F & \textbf{3.7532} & \textbf{4.0726} & \textbf{3.4907} & \textbf{3.8490} & \textbf{3.7181} & \textbf{3.9565} \\
\addlinespace
2 (213) & B & 3.7793 & 4.1725 & 3.4111 & 3.7941 & 3.7117 & 3.8712 \\
        & F & \textbf{3.8052} & \textbf{4.1746} & \textbf{3.4594} & \textbf{3.8248} & \textbf{3.7393} & \textbf{3.9048} \\
\addlinespace
3 (199) & B & 3.7482 & 4.0701 & 3.3887 & 3.7448 & 3.6495 & 3.8512 \\
        & F & \textbf{3.8210} & \textbf{4.1144} & \textbf{3.4931} & \textbf{3.8152} & \textbf{3.7531} & \textbf{3.8846} \\
\addlinespace
4 (206) & B & \textbf{3.8979} & \textbf{4.1255} & \textbf{3.6994} & 3.8953 & 3.6759 & 4.0867 \\
        & F & 3.8891 & 4.1235 & 3.6845 & \textbf{3.8998} & \textbf{3.6851} & \textbf{4.0872} \\
\bottomrule
\end{tabular}
\end{table}


\section{Discussion}
The interaction gain is front-loaded: minimal guidance produces most of the improvement, which matches the efficiency objective of the challenge. This behavior follows from the design of Phase C, which trains the network on scribbles placed inside its own largest residual errors, so that the first test-time correction delivers precisely the signal the model has learned to exploit, in line with prior evidence that training on self-generated errors benefits interactive models \cite{wang2018deepigeos,sofiiuk2022reviving}. The complementary tracer profiles, with higher Dice on FDG and higher DMM on PSMA, further suggest that a single corrective mechanism serves two distinct failure modes, volumetric refinement of large FDG lesions and detection of numerous small PSMA lesions, without any tracer-specific tuning. The complementarity between the best and final checkpoints observed on one fifth of the cases is the empirical basis of the ten-checkpoint ensemble, whose expected benefit over a five-checkpoint ensemble is small but consistently non-negative; because the additional checkpoints enter only through logit averaging, they alter neither the interaction protocol nor the compliance of the container with the per-invocation time limit. Remaining simplifications are deliberate choices in favor of robustness and reproducibility: binary (unsmoothed) interaction heatmaps introduce no encoding hyperparameters, stateless re-prediction mirrors the one-step-per-invocation execution model of the evaluation platform, and the threshold-based label convention of DeepPSMA is treated as tolerable training noise in exchange for broader multicenter and multitracer coverage.


\section{Conclusion}
A three-phase curriculum turns a standard residual-encoder nnU-Net into a scribble-adaptive segmenter that improves monotonically under corrective guidance. The curriculum requires no architectural modification beyond two additional input channels, so the mature self-configuring pipeline of the underlying framework carries over to the interactive setting intact. The observed gains are consistent across folds, tracers, and simulated scribble styles, which indicates that the learned corrective behavior is a property of the training regime rather than of a particular annotation pattern. 

\begin{credits}
\subsubsection{\discintname}
The authors have no competing interests to declare that are relevant to the content of this article.
\subsubsection{\ackname} 
The authors use Claude Fable 5~\cite{anthropic2026claudefable5} to assist with correcting grammar and refining readability. 
\end{credits}
\clearpage

\bibliographystyle{splncs04}
\bibliography{refs}

\clearpage
\appendix

\section{Training Curves}
\begin{figure}[!ht]
\centering
\includegraphics[height=0.80\textheight,keepaspectratio]{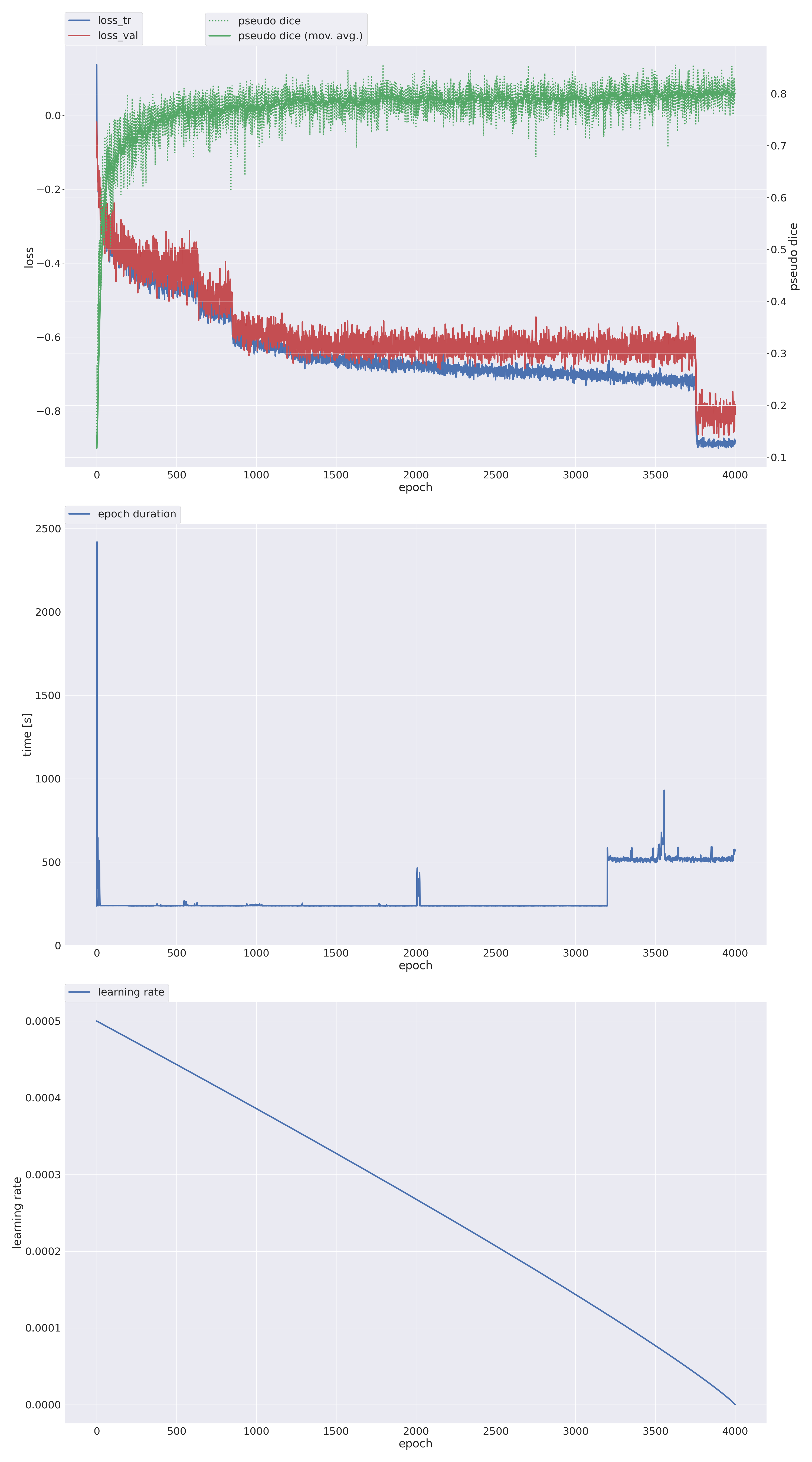}
\caption{Training progress of fold 0.}
\label{fig:tr0}
\end{figure}

\begin{figure}[p]
\centering
\includegraphics[height=0.95\textheight,keepaspectratio]{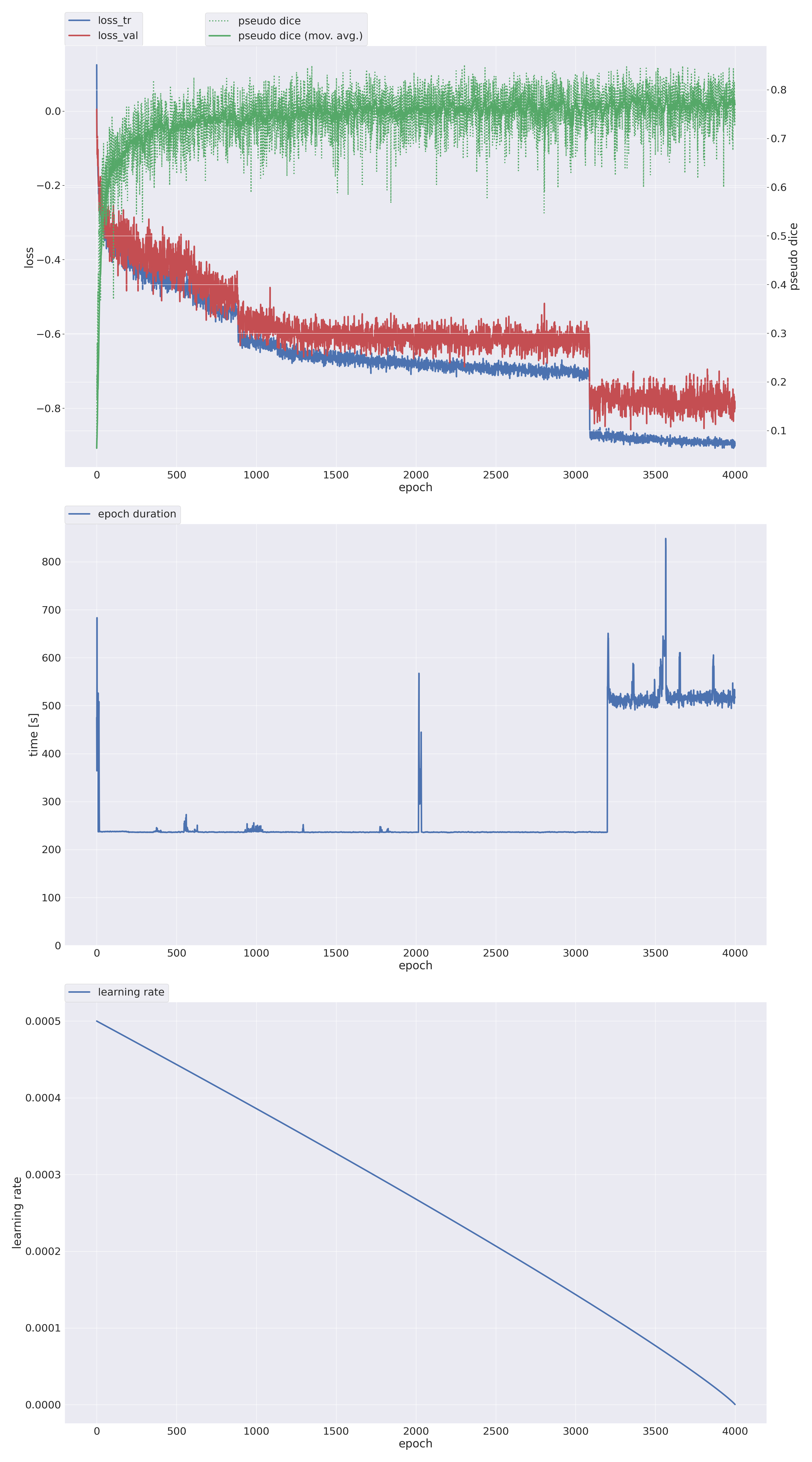}
\caption{Training progress of fold 1.}
\label{fig:tr1}
\end{figure}

\begin{figure}[p]
\centering
\includegraphics[height=0.95\textheight,keepaspectratio]{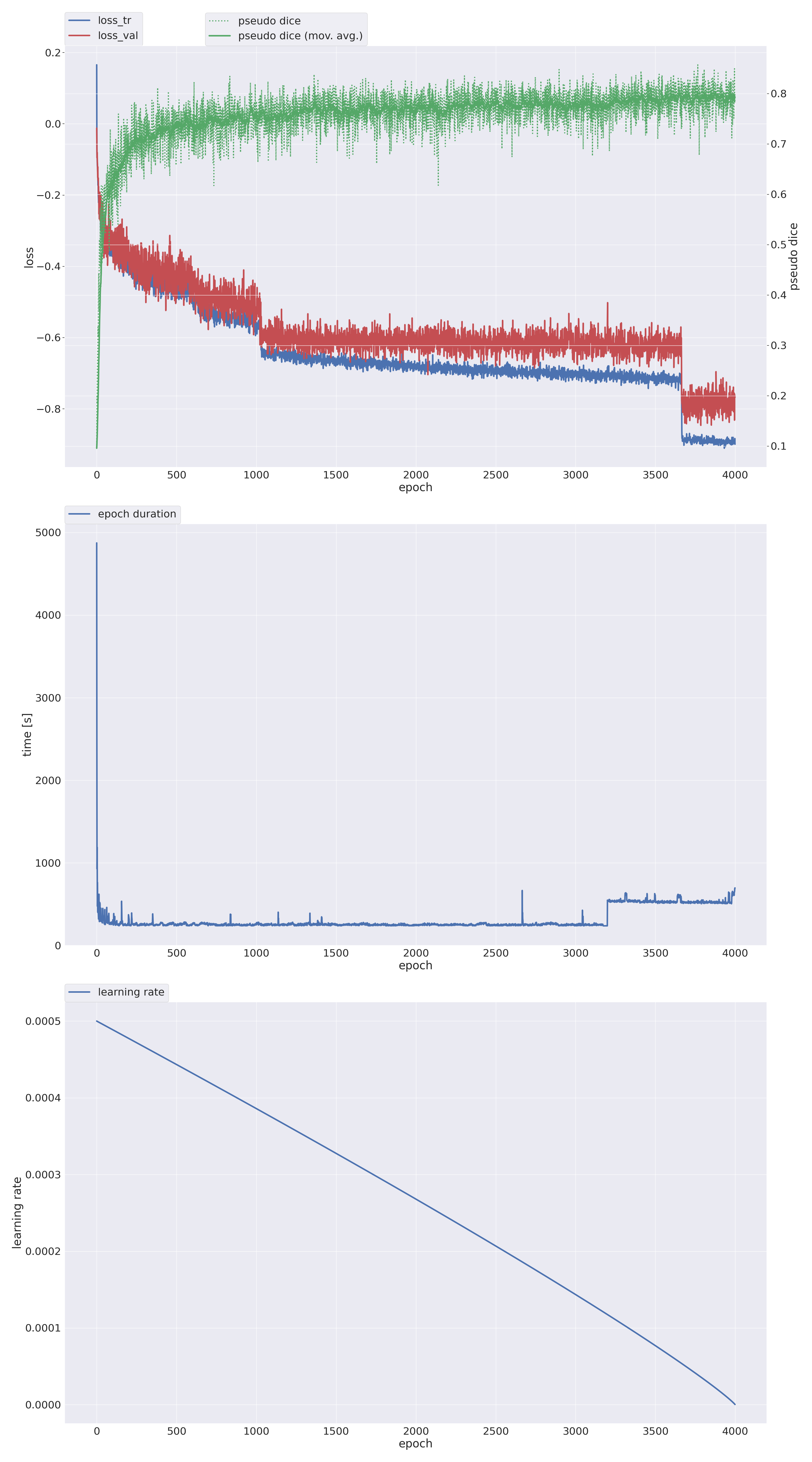}
\caption{Training progress of fold 2.}
\label{fig:tr2}
\end{figure}

\begin{figure}[p]
\centering
\includegraphics[height=0.95\textheight,keepaspectratio]{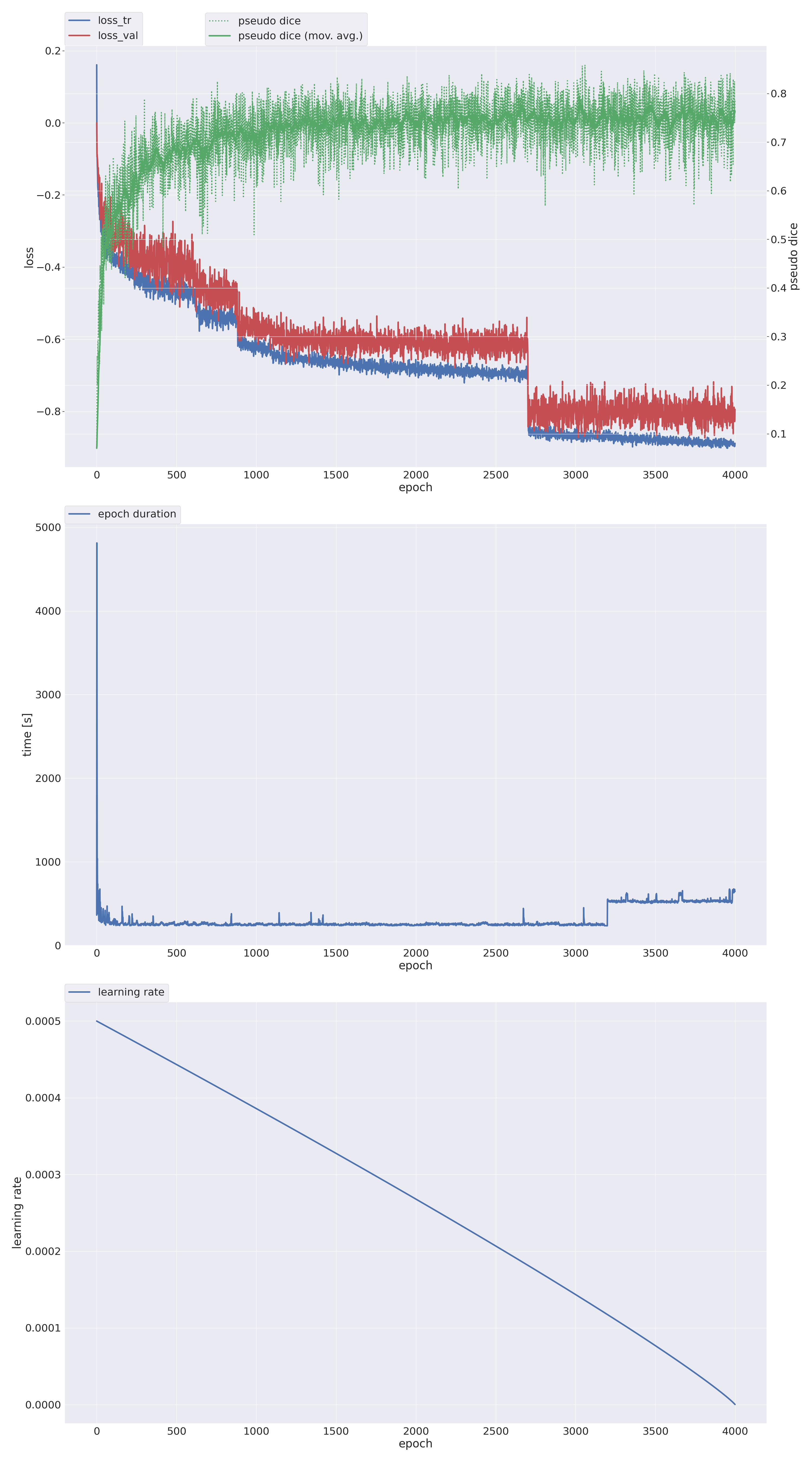}
\caption{Training progress of fold 3.}
\label{fig:tr3}
\end{figure}

\begin{figure}[p]
\centering
\includegraphics[height=0.95\textheight,keepaspectratio]{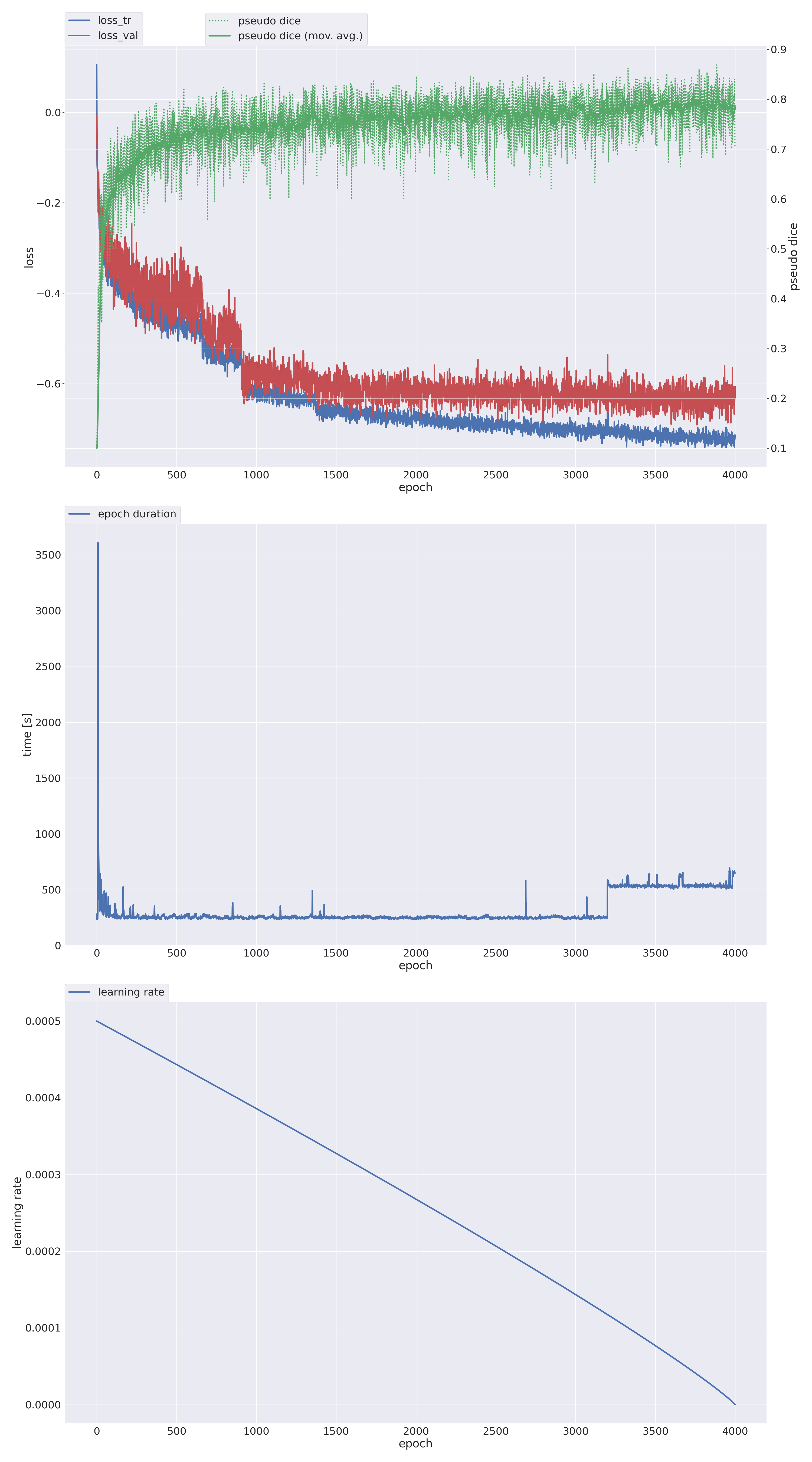}
\caption{Training progress of fold 4.}
\label{fig:tr4}
\end{figure}

Figures~\ref{fig:tr0} to \ref{fig:tr4} illustrate nnU-Net's training progress of all five folds. The epoch-duration step at epoch 3200 marks the onset of Phase C, whose additional gradient-free forward pass roughly doubles the duration of every training epoch.

\clearpage 

\clearpage

\clearpage

\clearpage

\end{document}